\documentclass{article}

     \PassOptionsToPackage{numbers, compress}{natbib}

    \usepackage[preprint]{neurips_2020}

\usepackage[utf8]{inputenc} 
\usepackage[T1]{fontenc}    
\usepackage{hyperref}       
\usepackage{url}            
\usepackage{booktabs}       
\usepackage{amsfonts}       
\usepackage{nicefrac}       
\usepackage{microtype}      
\usepackage{graphicx}
\usepackage{xspace}
\usepackage{amsmath}
\usepackage{subcaption}

\makeatletter
\DeclareRobustCommand\onedot{\futurelet\@let@token\@onedot}
\def\@onedot{\ifx\@let@token.\else.\null\fi\xspace}
\def\eg{\emph{e.g}\onedot} 
\def\ie{\emph{i.e}\onedot} 
 
\def\etc{\emph{etc}\onedot} \def\vs{\emph{vs}\onedot}

\makeatother

\title{Image-Graph-Image Translation via Auto-Encoding}

\author{%
  Chenyang Lu and Gijs Dubbelman \\
  Department of Electrical Engineering\\
  Eindhoven University of Technology\\
  Eindhoven, The Netherlands \\
  \texttt{\{c.lu.2, g.dubbelman\}@tue.nl} \\
}

\begin{document}

\maketitle

\begin{abstract}
This work presents the first convolutional neural network that learns an image-to-graph translation task without needing external supervision. Obtaining graph representations of image content, where objects are represented as nodes and their relationships as edges, is an important task in scene understanding. Current approaches follow a fully-supervised approach thereby requiring meticulous annotations. To overcome this, we are the first to present a self-supervised approach based on a fully-differentiable auto-encoder in which the bottleneck encodes the graph's nodes and edges. This self-supervised approach can currently encode simple line drawings into graphs and obtains comparable results to a fully-supervised baseline in terms of F1 score on triplet matching. Besides these promising results, we provide several directions for future research on how our approach can be extended to cover more complex imagery.
\end{abstract}

\section{Introduction}

Scene understanding from sensory measurements is an essential task for intelligent agents, and has been widely investigated in recent years. The advances of deep learning brought scene understanding into a new era, especially for semantic image understanding tasks starting from object detection \cite{girshick_rich_2014} to panoptic segmentation \cite{kirillov_panoptic_2018}. It can be argued that the aforementioned tasks provide the scene representations majorly in the context of pixels, \eg bounding boxes, per-pixel semantic labels, \etc, which are not the ideal data structures for high level reasoning and decision making by intelligent agents. An alternative option for representing geometric and semantic image content is a graph. Compared to pixel-based representations, a graph, \eg OpenStreetMap \cite{haklay_openstreetmap:_2008} or a scene graph as in \cite{johnson_image_2015}, is a much better data structure to store scene information for high level reasoning and decision making and, at the same time, being an efficient representation in terms of memory and compute requirements. While the translation from image content (like bounding boxes \etc.) into a graph can be performed with hand-designed logic, such approaches are highly task-dependent and thus not scalable. Therefore, similar to our work, significant contemporary research is devoted into methods that can learn to translate image content in a (scene) graph.

Practically all current state-of-the-art methods for this image-to-graph task, like \cite{xu_scene_2017, herzig_mapping_2018, zellers_neural_2018, yang_graph_2018, li_factorizable_2018, qi_attentive_2019} and as discussed in Section \ref{sect_related_work}, build on top of an object detection framework and rely on meticulously annotated datasets that contain the relationships between objects, as the ground truth for fully-supervised training. Although large scale datasets for this task are available \cite{krishna_visual_2017}, the time required for the needed annotations is so significant that also these fully-supervised learning-based approaches do not scale well to task for which the required ground truth is not yet available. Therefore, we set the basic research goal to be able to learn the image-to-graph translation task in a self-supervised manner by extending the methodology of auto-encoding. While the current capability of our approach is limited to simple line drawings, we believe that it holds the key to developing image-to-graph methods that scale over many different scene understanding tasks and thus do not rely on hand-crafted logic or meticulously annotated datasets.

The details of our graph auto-encoding approach are provided in Section \ref{sect_method}. At its core, it learns to translate image content into a graph that is represented by node and adjacency matrices. This learning is self-supervised using a fully-differential auto-encoder in which the decoder consists of several carefully designed neural networks and the decoder uses techniques from differentiable image drawing. The image-to-graph-to-image task is then learned in an auto-encoder fashion by minimizing the pixel loss between the input image and the image generated by the decoder. During inference, only the encoder is used, although the learning procedure can, in theory, also be used on-line to fine-tune the obtained graph.   

To demonstrate our approach, we use a synthetic dataset that contains several line drawings of simple shapes, each of which represents a graph with nodes and edges. The details of the dataset and its corresponding experimental results are presented in Section \ref{sect_method} and Section \ref{sect_exp}. Although the demonstrated experiments are still at an early stage and further work must be carried out to generalize to large scale real datasets like Visual Genome \cite{krishna_visual_2017}, we believe our findings are worth to be shared with the community. Our contributions include:
\begin{itemize}
	\item the, to the best of our knowledge, first neural network that can learn an image-to-graph translation task by only using self-supervision; 
	\item a comparison of our method to an fully-supervised equivalent baseline, which shows our approach exhibits comparable performance in terms of F1-score of triples matching;
	\item several ablation studies that examine and reveal key properties of our method.
\end{itemize}

\begin{figure}
	\centering
	\includegraphics[width=0.8\linewidth]{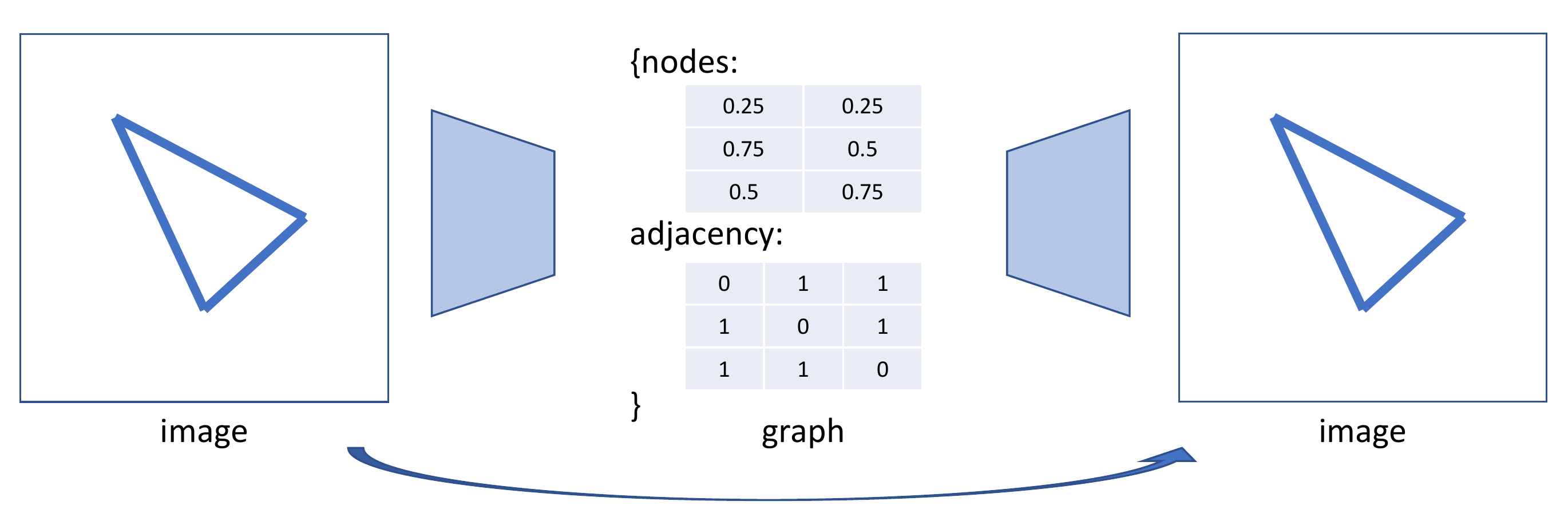}
	\caption{The overview of the proposed approach. The auto-encoder is able to extract the canonical graph information from an input image at the bottleneck, with only the image-level self-reconstruction loss.}
\end{figure}

\section{Related work}
\label{sect_related_work}

\paragraph{Scene graph generation}
The task of scene graph generation, as pioneered by \citet{johnson_image_2015}, is to estimate a graph structure representing the relevant information of the scene from sensory measurements, In the graph, the objects are nodes and their relationships are modeled as edges. Various approaches are proposed in recent years, enabled by the release of the Visual Genome dataset \cite{krishna_visual_2017} with massive manual annotations. Most of the state-of-the-art approaches \cite{xu_scene_2017, herzig_mapping_2018, zellers_neural_2018, yang_graph_2018, li_factorizable_2018, qi_attentive_2019} are formulated on top of bounding box-based detection frameworks, while using advances in graph convolutional networks (GCN) \cite{yang_graph_2018} and graph attention mechanisms \cite{qi_attentive_2019} to construct the scene graph. In \citet{newell_pixels_2018} it is proposed to generate heat-maps for both objects (nodes) and relations (edges) simultaneously for further graph construction using associative embeddings. Recently, scene graph generation is also extended into 3D \cite{armeni_3d_2019} and applied to input data other than images, such as point clouds. \cite{wald_learning_2020}. 

It is important to note that these mentioned state-of-the-art methods are, unlike our method, highly dependent on massive annotated ground truth for training. While annotating for object detection is rather straightforward, \ie, an object is present or not, annotating the relationships between objects is much more involved. It does not only depend on the image context but also on the task what relationships are relevant to be annotated and it is practically impossible to annotate all possible relationships between objects for all possible tasks. This is also why the main performance criteria for image-to-graph translation is currently based on recall but not on precision, \ie, methods typically estimate many redundant relationships that are not present in the ground truth annotations but that, in certain cases, can be considered as being correct. While the Visual Genome dataset \cite{krishna_visual_2017} is extensive it surely does not cover the need for ground truth annotations for all possible tasks in computer vision that can benefit from obtaining graph representations from images. As such, to make available image-to-graph translation methods that can effortlessly scale to new tasks, we set the basic research goal to investigate and develop methods that require as little ground truth annotations as possible and preferably none.

\paragraph{Unsupervised scene representation learning}
In general, to reduce the dependency on expensive manual annotations, unsupervised approaches have been widely investigated. The learned representations include individual objects and their information, physical factors, \etc. 

In \citet{burgess_monet_2019} a variational auto-encoder is used and trained together with a recurrent attention network decomposing the objects in an image. This model is fully differentiable and can be trained with self-reconstruction loss in an end-to-end manner. More recently, the work in \citet{yang_learning_2020} went one step further to decompose the image into objects and enable object manipulation without requiring object-level annotations. Also, in \citet{wu_unsupervised_2020} the proposed encoder-decoder framework factors each input image into depth, albedo, viewpoint, and illumination, using only unsupervised reconstruction loss. This is achieved by interpreting the prior knowledge of symmetric structure in the design of the network's information flow.

The mentioned approaches achieve unsupervised learning of scene representation mainly by carefully designing the framework using human prior knowledge and training the models in an auto-encoding fashion with reconstruction supervision. Our approach has the same philosophy, but is more aggressive in the representation-shifting that is performed. In our approach, the self-learned representation (graph) is further away from the input representation (image) in terms of the format and the level of semantic information, than in other works where the self-learned representations are image-level objects.

\paragraph{Differentiable image synthesis}
An important part of our approach is the ability to decode a graph into an image. Conventional computer graphics (CG) methods generate realistic images by approximating the physical processes of light, which are typically hard-coded and non-differentiable. As such, they are not designed to be integrated with deep learning methods. To overcome this, differentiable renderers \cite{loper_opendr_2014, kato_neural_2018} are proposed to mimic the classical CG process with differentiable operations, which enables various tasks of rendering in combination with using deep neural networks. These renderers are mainly conventional geometry-based approaches with novel differentiable extensions. Thus, they are not able to have the abstract semantic understanding of the scene.

Significant research is carried out on image synthesis using CNN-based generative networks, such as image-to-image translation \cite{isola_image--image_2017, zhu_unpaired_2017}. CNN-based approaches are able to capture the complicated high level information of the scene to be generated, \eg various background layouts, object classes and their attributions, \etc. However, these approaches are mostly data-driven and lack the explainability of the representations inside the CNNs. 

In contrast, scene graphs are better able to provide high level explainable representations than conventional tensors or feature maps.
More close to our research, scene graphs are also used in the image synthesis tasks as input \cite{johnson_image_2018}. Furthermore, image manipulation is also realized by interactively modifying the scene graph extracted from the input image in the bottleneck of an encoder-decoder network \cite{dhamo_semantic_2020}. However, note that in \cite{dhamo_semantic_2020} their explainable scene graph representation is first massively annotated by humans and then further learned by the model, unlike our approach that employs self-supervised learning and does not require any human labeling effort.

\section{Methodology}
\label{sect_method}

\subsection{Problem definition}

We start with the conventional problem definition of scene graph generation task and then discuss the similarities and differences in our setting. 

Given an input image $I$, we assume that there exists a scene graph $G = (V, E)$ with $V$ being a set of nodes corresponding to localized object regions in image $I$, and $E$ being the edges representing the relationships between nodes $V$. Note that each element $v_i$ and $e_{ij}$ in $V$ and $E$ could have one of multiple semantic labels. Thus, the scene graph generation can be formulated as the mapping $f: I \mapsto G$. Most approaches \cite{xu_scene_2017, herzig_mapping_2018, zellers_neural_2018, yang_graph_2018, li_factorizable_2018, qi_attentive_2019} factorize it mainly into two sequential sub-mappings, \ie $f = f_E \cdot f_V$, where
\begin{equation}
\begin{array}{l}
f_V: I \mapsto V \\
f_E: (I, V) \mapsto E.
\end{array}
\end{equation}

$f_V:I \mapsto V$ is often accomplished by object detection frameworks, and $f_E: (I, V) \mapsto E$ is the relationship classification. Both of these mapping processes are usually performed by learnable neural networks. In the conventional fully-supervised setting, for each training sample the ground truth nodes $\bar{V}$ and edges $\bar{E}$ are provided such that the neural networks performing the mapping $f: I \mapsto G$ can be learned.

If the ground truth $\bar{V}$ and $\bar{E}$ are not available, the previous approaches cannot be applied, as they are discriminative models trained in a supervised manner. On the contrary, we opt to extend the task of scene graph generation to image-graph-image translation by using the concept of auto-encoding. The encoder $f: I \mapsto G$ is expected to perform the same task as the scene graph generation \cite{johnson_image_2015}, only in our case, the supervision is not provided. It is also composed of two sub-modules, \ie, $f_V: I \mapsto V$ for node prediction and $f_E: (I, V) \mapsto E$ for edge prediction. The decoder $g: G \mapsto I$ takes the intermediate graph information as input, and re-generates the input image in a fully differentiable manner. By placing the encoder and decoder behind each other, techniques from auto-encoding can be used to learn the graph information.

Formally, our goal is to have two mapping functions $f$ and $g$,
\begin{equation}
\begin{array}{l}
\tilde{G} = f(I) = f_E \cdot f_V (I) \\
\tilde{I} = g(\tilde{G}), \\
\end{array}
\end{equation}
such that
\begin{equation}
\begin{array}{l}
\tilde{G} \in \mathcal{G}, \\
\tilde{I} \in \mathcal{I}, \\
f,g = \underset{f,g}{\arg\min}\ \mathcal{L}(I, \tilde{I}), \\
\end{array}
\end{equation}
where $\mathcal{G}$ and $\mathcal{I}$ are the collection of all possible graphs and images respectively, and $\mathcal{L}(\cdot, \cdot)$ is the similarity measure of two images. 

In the following sub-sections, we will detail the design and training protocol of the proposed approach. We take a simple toy task with a synthetic dataset as an example to illustrate the idea. The dataset contains a bunch of undirected graphs which are represented as images. The nodes in the graphs are the end-point of the visible edges and the edges are defined as the binary connectivity between two nodes. Note that ideally, the task and dataset can be extended with more complicated definitions, however as the first step, we opt to simplify the task and focus on the basic methodology. The details of the dataset will be discussed in Section~\ref{sect_exp}.

\subsection{Network}
\label{sect_method_net}

\begin{figure}
	\centering	\includegraphics[width=\linewidth]{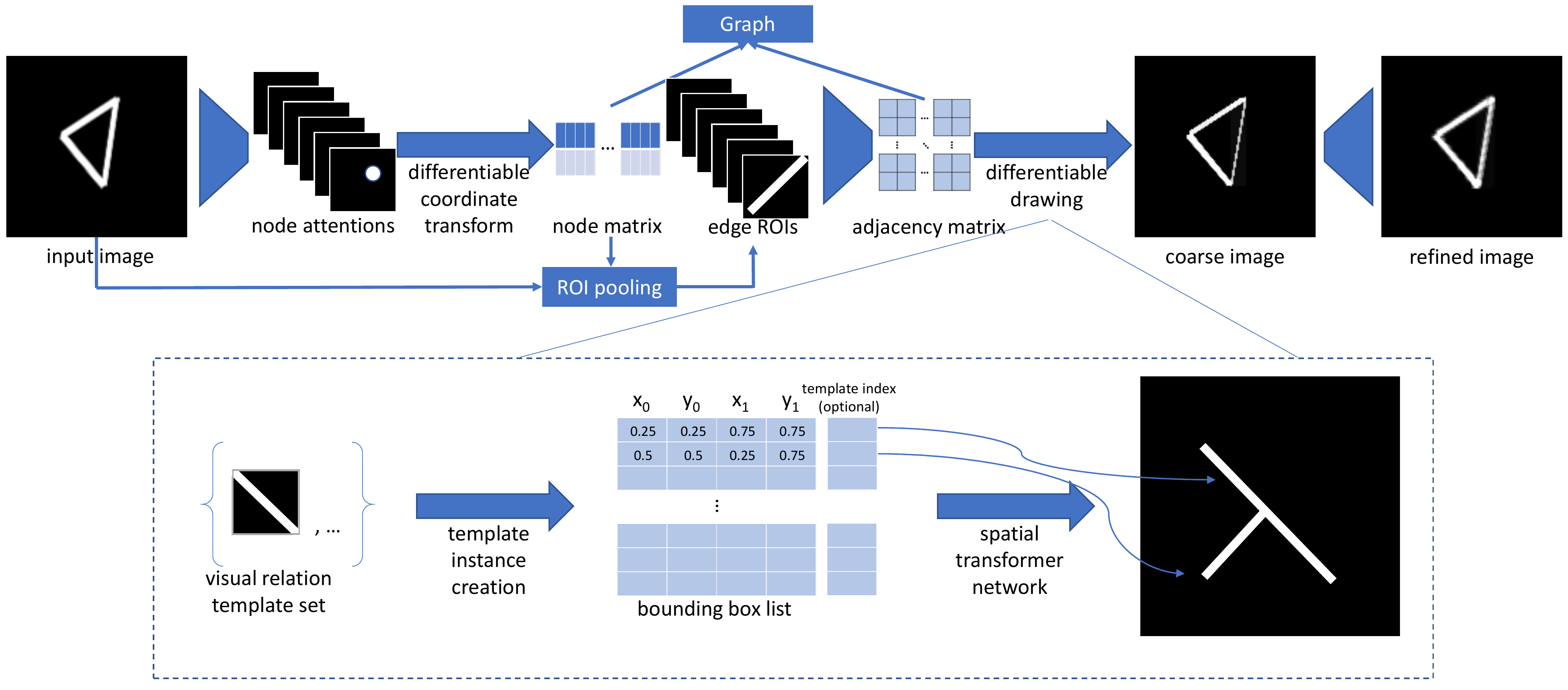}
	\caption{The overall framework of the proposed approach, with the upper part being the detailed pipeline and the lower part being the differentiable drawing module. The entire framework is trained end-to-end and using only self-reconstruction supervision. Please See Section \ref{sect_method_net} for more details.}
	\label{fig_network}
\end{figure}

The overall design of the auto-encoder for image-graph-image translation task is visualized in Figure~\ref{fig_network}. It mainly consists of three differentiable modules, i.e. node prediction, edge classification and differentiable image synthesis. The three modules are sequentially connected and trained in an end-to-end manner using image self-reconstruction loss.

\subsubsection{Encoder: from image to graph}

As discussed previously, the encoder part of the network is to predict the graph information from the input image, which is formalized as the process $f = f_E \cdot f_V :I \mapsto G$. The encoder is composed of three sequential parts, namely node attention, coordinates transformation, and relation classification. The output of the encoder, \ie the bottleneck of the auto-encoder, is the graph represented by two matrices, i.e. the node position matrix and the adjacency matrix.

\paragraph{Node attention} 

At the early stage, the convolutional layers from ResNet-50 \cite{he_deep_2016} are implemented as the feature extractor. Since our task is much simpler, we take features from the second residual block and reduce the output channels of two blocks to 128 and 64, respectively. To reduce the size of the extracted features, two max-pooling layers with kernel size and stride being 2 are also applied before each block. On top of the residual blocks we have two extra convolutional layers to predicts the node attention maps $M_{att}$, with the spacial size reduced by the factor of 4 due to the previous pooling operations, and with the number of channels being the pre-defined maximum number of nodes $N_{max}$. Each channel in node attention maps $M_{att}$ is passed through a 2-D softmax layer, such that the summation of all the elements strictly equals to 1, which is essential for the later computation of node coordinates. Note that here we expect the attention channels are able to provide channel-wise heat maps indicating the detected nodes. However, no supervision or extra information is supplied to the network, and the network is able to learn from the final reconstruction loss by gradient back-propagation.

\paragraph{Coordinates transformation} 

Being different from other detection frameworks \cite{girshick_rich_2014, girshick_fast_2015, ren_faster_2017}, in our setting the coordinates of nodes are not provided as the ground truth. Thus the differentiability is crucial at this stage, since without which the gradient back-propagation cannot be applied and the node attention module cannot learn anything. However, it is non-trivial to transform the positional information in the heat maps to numerical coordinates in a differentiable manner. Here we use differentiable spatial to numerical transform (DSNT) \cite{nibali_numerical_2018} to perform the transformation. Due to the usage of the previously mentioned 2-D softmax layer, for each channel, the heat map can be seen as a 2-D probabilistic distribution of a certain node. Thus, one can create two fixed template maps containing the numerical coordinates value for each pixel, in horizontal and vertical directions, respectively. With element-wise production and summation, the numerical coordinates for each node can be computed. At this stage, there exist no trainable parameters, and the computation, which is detailed in \cite{nibali_numerical_2018}, is fully deterministic and differentiable.

\paragraph{Edge prediction}

Once the coordinates of the nodes are computed, a set of regions of interest (ROIs) for edges can be constructed by combining arbitrary two node coordinates as a bounding box. Then ROI pooling \cite{ren_faster_2017} is applied on the input image and creates a batch of local image patches of size $N_{max}^2 \times 1 \times 16 \times 16$ for further edge prediction by a edge classifier. The classifier consists of two convolution layers (kernel size 3 and output channels 32 and 16, respectively) with Batch Normalization, ReLU activation and max-pooling. Then, the features are flattened and processed by two fully-connected layers with ReLU and Sigmoid activations, respectively.

Please note that in our toy task the edge represents the connectivity for the sake of simplicity, while it can be extended with semantic information with little extra effort. The predicted edge connectivities, together with the previously computed nodes, make up the output of the encoder, \ie a graph with nodes position matrix and adjacency matrix.

\subsubsection{Decoder: from graph to image}

The task of the decoder $g: G \mapsto I$ is to reconstruct the input image given the graph information provided from the encoder. The key is to ensure the reconstruction flow is fully differentiable such that the whole pipeline can be trained together end-to-end. Inspired by \cite{johnson_image_2018}, we consider a template-based image synthesis approach, which is referred to as \textit{differentiable drawing}. 

We create a template set that contains all the pre-defined relationships that could exist in the dataset. For each forward-pass of the network, the differentiable drawing module walks through all the edges in the predicted adjacency matrix, and for the edges that need to be visualized in the image, spatial transformer network \cite{jaderberg_spatial_2015} is implemented to copy and draw the corresponding edge template onto the canvas. The differentiable drawing module is able to create a coarse reconstructed image $\tilde{I}_{\text{coarses}}$ on a blank canvas. This process can be written as $g_{\text{coarse}}: G \mapsto \tilde{I}_{\text{coarse}}$. On top of the coarse image $\tilde{I}_{\text{coarses}}$, a refinement network $g_{\text{refine}}: \tilde{I}_{\text{coarse}} \mapsto \tilde{I}_{\text{refine}}$ is applied for further post-processing, which does not change the structural information but modifies and eliminates the textures and styles difference between the created coarse image and real input image. In our implementation, the refinement network contains three convolution layers with kernel size 3, intermediate channel size 16, and PReLU activations between every two layers.


\subsection{Training}

The entire framework is trained end-to-end without supervisions that require manual annotations. To achieve this goal, we apply two losses during training: the main image reconstruction loss for auto-encoding and the auxiliary node attention loss.

As the main image reconstruction loss, a structural similarity index measure (SSIM) \cite{wang_image_2004} is used, which encourages the decoder to reconstruct the input image and thus encourages the encoder to understand the image in terms of graphs in an implicit manner. We apply SSIM losses with multi-scale setting on the refined images in the experiments, which is formulated as 
\begin{equation}
	\mathcal{L}_{\text{main}} = \text{MS-SSIM}(\tilde{I}_{\text{refine}}, I)
\end{equation} 
where $\tilde{I}_{\text{refine}}$ and $I$ are the reconstructed refined image and input image, respectively.

The node attention module is expected to predict the node positional attention maps from the input image solely relying on the gradient information from the reconstruction loss. Due to the lack of explicit supervision, the attention module occasionally tends to predict some nodes multiple times while ignoring other nodes. To overcome this behavior and stabilize the node attention training, we apply the \textit{auxiliary loss} $\mathcal{L}_{\text{aux}}$. The idea is to penalize the overlay behavior between the predicted node attentions and encourage the attention module to discover as many different nodes as possible. On the other hand, since the pre-defined maximum number of nodes (attention channels) is fixed and is always larger than the real value, the overlay is expected to happen. We propose a penalty loss conditioned on the similarity measure for each sample, \ie, the penalty is applied to the samples only if the reconstruction is not good enough. The auxiliary loss can be written as 
\begin{equation}
\mathcal{L}_{\text{aux}} = \left\{
	\begin{array}{ll}
	\text{mean}([\text{ReLU}(\sum_{i}^{N_{\text{max}}}M_{att}[i] - 1)]^2),& \text{if MS\_SSIM < mean MS\_SSIM} \\ 
	0, & \text{if MS\_SSIM > mean MS\_SSIM}
	\end{array}\right.
\end{equation}
where $M_{att}$ is the normalized predicted node attention maps with the maximum cell for each channel being 1. MS\_SSIM and mean MS\_SSIM are the similarity measures of each sample and the mean measure of the entire batch during training. All the operations are element-wise, and we skip the height and weight index for attention maps for the sake of simplicity.

Formally, the overall loss can be written as
\begin{equation}
\mathcal{L} = \mathcal{L}_{\text{main}} + \lambda \cdot \mathcal{L}_{\text{aux}}
\end{equation}
where $\lambda$ is the weight for loss balancing and is empirically set to 1 in our experiments.

\section{Experiments}
\label{sect_exp}
We perform the following experiments to demonstrate and verify the proposed method: 1) we compare the results of our unsupervised approach and a supervised baseline, qualitatively and quantitatively; 2) we perform an ablation study of our approach using different maximum number nodes; and 3) we also study the effect of different image reconstruction losses.

\begin{figure}
	\centering
	\begin{subfigure}[t]{.11\textwidth}
	\centering
	\includegraphics[width=\linewidth]{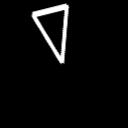}\\
	\includegraphics[width=\linewidth]{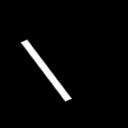}\\
	\includegraphics[width=\linewidth]{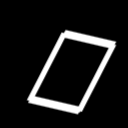}\\
	\caption{\centering input image}
	\end{subfigure}
	\begin{subfigure}[t]{.11\textwidth}
	\centering
	\includegraphics[height=\linewidth]{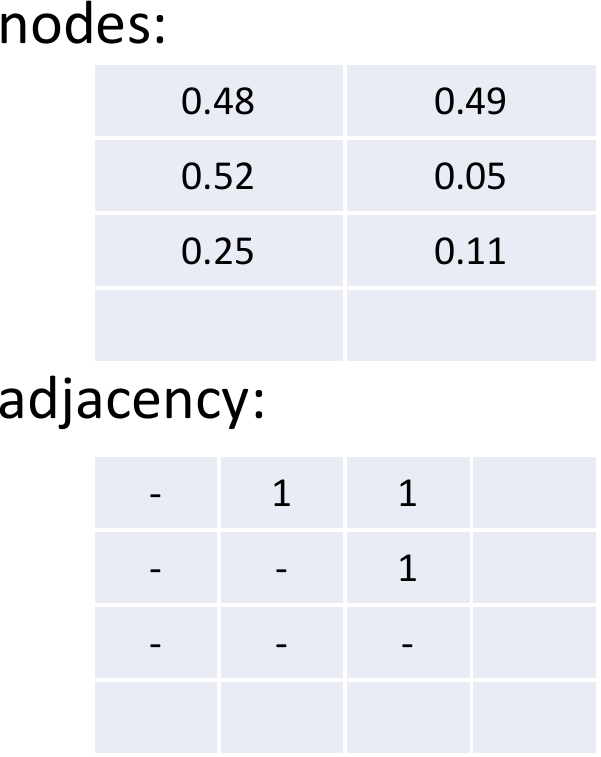}\\
	\includegraphics[height=\linewidth]{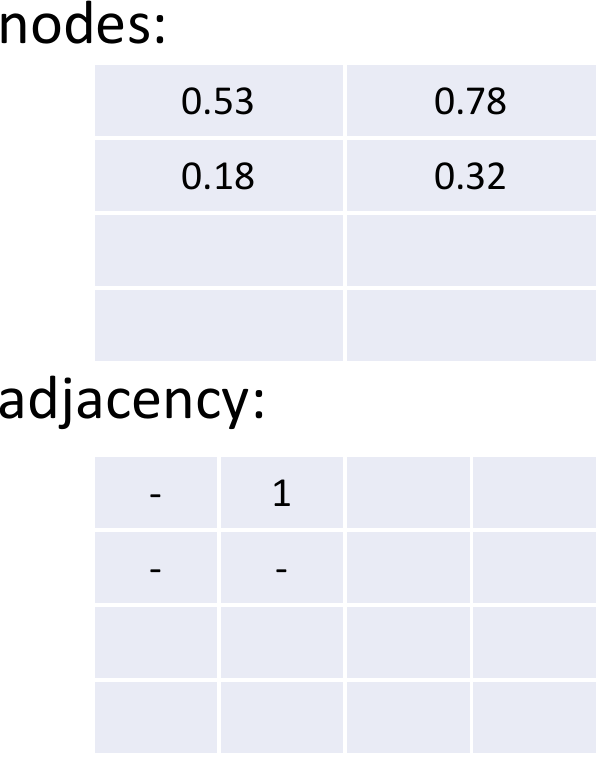}\\
	\includegraphics[height=\linewidth]{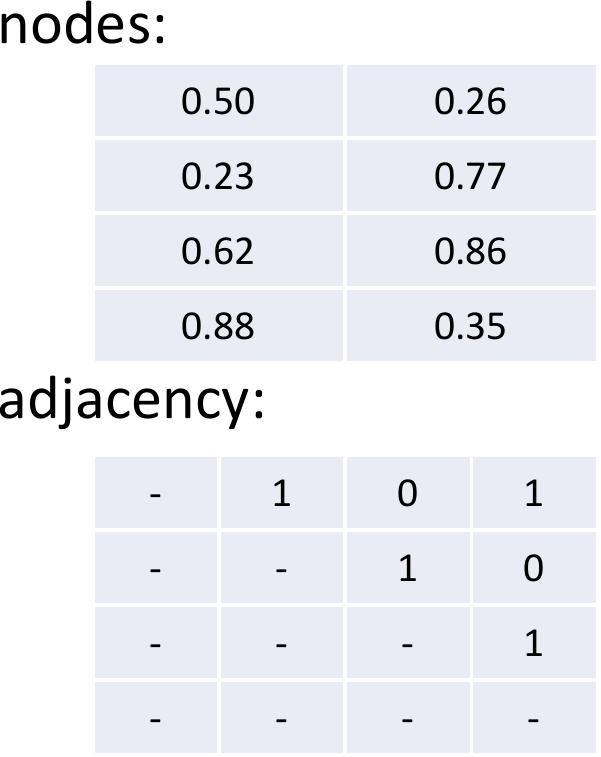}\\
	\caption{\centering ground truth graph}
	\end{subfigure}
	\begin{subfigure}[t]{.11\textwidth}
	\centering
	\includegraphics[width=\linewidth]{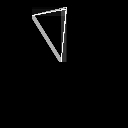} \\
	\includegraphics[width=\linewidth]{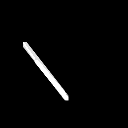} \\
	\includegraphics[width=\linewidth]{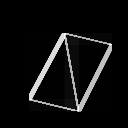} \\
	\caption{\centering coarse image}
	\end{subfigure}    
	\begin{subfigure}[t]{.11\textwidth}
	\centering
	\includegraphics[width=\linewidth]{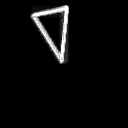} \\
	\includegraphics[width=\linewidth]{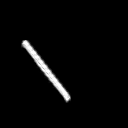} \\
	\includegraphics[width=\linewidth]{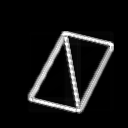} \\
	\caption{\centering refined image}
	\end{subfigure}
	\begin{subfigure}[t]{.11\textwidth}
	\centering
	\includegraphics[width=\linewidth]{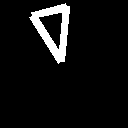}\\
	\includegraphics[width=\linewidth]{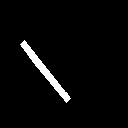}\\
	\includegraphics[width=\linewidth]{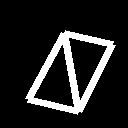}\\
	\caption{\centering ours prediction (image)}
	\end{subfigure}
	\begin{subfigure}[t]{.11\textwidth}
	\centering
	\includegraphics[height=\linewidth]{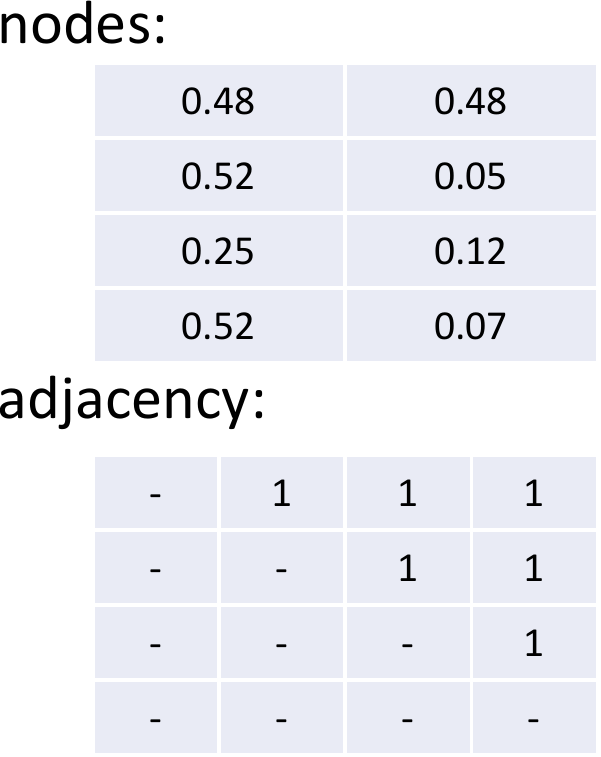}\\
	\includegraphics[height=\linewidth]{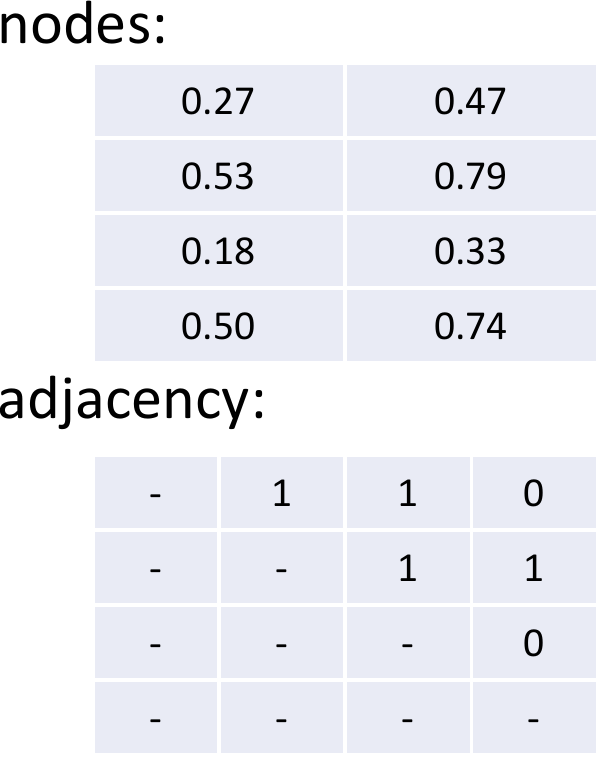}\\
	\includegraphics[height=\linewidth]{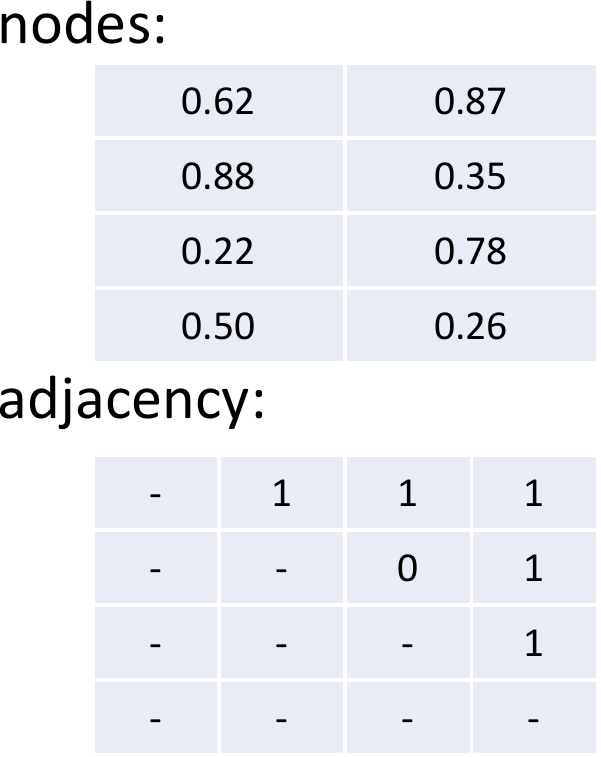}\\
	\caption{\centering ours prediction (graph)}
	\end{subfigure}
	\begin{subfigure}[t]{.11\textwidth}
	\centering
	\includegraphics[width=\linewidth]{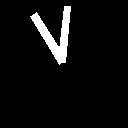}\\
	\includegraphics[width=\linewidth]{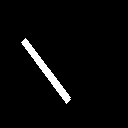}\\
	\includegraphics[width=\linewidth]{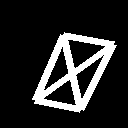}\\
	\caption{\centering baseline prediction (image)}
	\end{subfigure}
	\begin{subfigure}[t]{.11\textwidth}
	\centering
	\includegraphics[height=\linewidth]{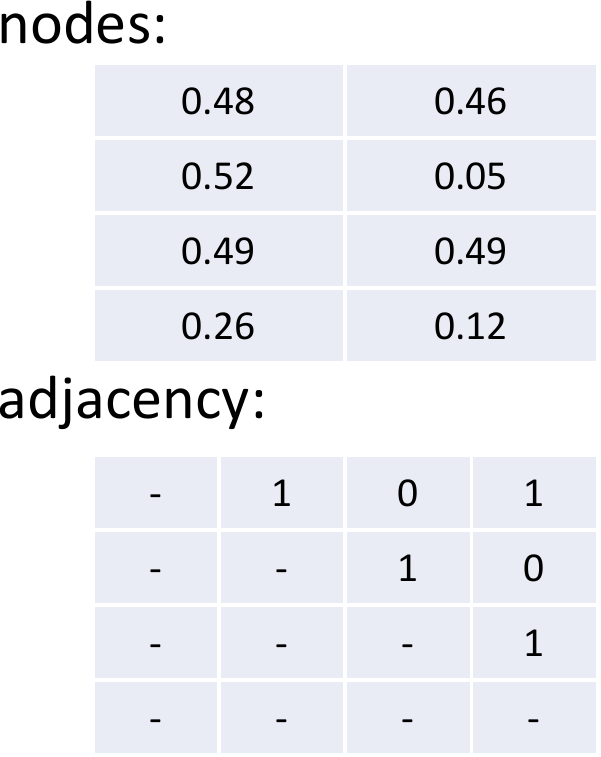}\\
	\includegraphics[height=\linewidth]{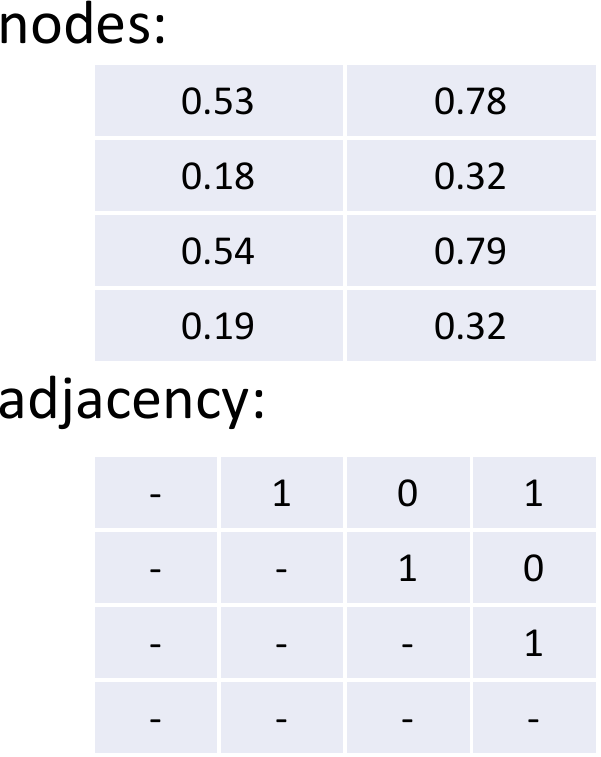}\\
	\includegraphics[height=\linewidth]{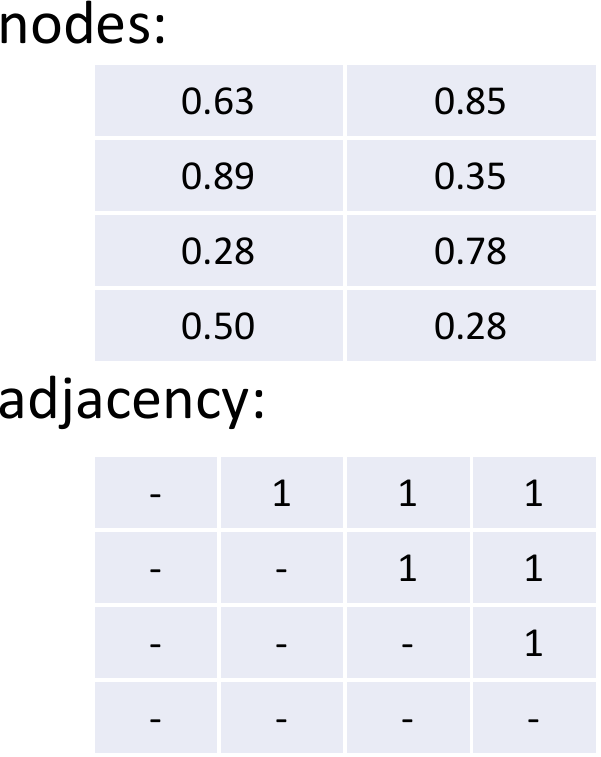}\\
	\caption{\centering baseline prediction (graph)}
	\end{subfigure}
	\caption{The qualitative results of our approach. (a)-(b) are the inputs image and the corresponding graphs. (c)-(f) are the predictions from our approach, in which (c)-(d) are the output images from the network and (e)-(f) are the graph predictions in image and matrix formats. (g)-(h) are the graph predictions from the baseline, in image and matrix formats, respectively. }
	\label{fig_qualitative}
\end{figure}

\paragraph{Datasets:}
We use a synthetic dataset, called Simple Shape dataset, to demonstrate the proposed idea. The dataset contains 50k images with each presenting one of the three simple shapes including line, triangle, and rectangle. Each shape is processed with random affine transformation (including scale, rotation, shear, and translation) and is drawn on an empty black canvas with the size being $128\times128$ pixels. Please see Fig~\ref{fig_qualitative} for some visualized samples. Thus, for each image, the number of nodes varies from 2 to 4, which is why the maximum number of nodes is set as 4 in the main experiments.

\paragraph{Baseline:} 

Since there is no existing previous method that performs the same task, \ie generating graphs from images without any external supervision, we provide a baseline that shares a similar design of the network, but has access to the ground truth graph information. To have a fair comparison, the capacity of the baseline network, is aligned with the design of the proposed network as much as possible, \eg the overall structure, number of layers, and channels are set as similar as possible. Since the ground truth is provided, the decoder is no longer needed for the baseline, and we directly apply two supervisions on the graph outputs of the encoder. First, the supervision for the node attention channels is provided by creating a ground truth heat map with Gaussian kernels at the node ground truth locations. As the node attentions can have random orders, we perform summation across the channels and compute the mean squire pixel error between the single channel attention and the ground truth heat map. Second, for each step, we perform the node matching between the prediction and the ground truth, to re-generate the temporary ground truth adjacency matrix aligned with the correct temporary node order. This allows for a cross-entropy classification loss to be computed. With the aforementioned two supervisions, \ie, pixel-wise heat map error and cross-entropy classification error, the graph can be learned in a fully-supervised manner.

Note that the baseline is intended to formulate a relatively fair comparison with our approach. Thus, it is preferred that the two models have similar capacities, which is why we apply minimal modifications to our model. One could carefully re-design the nodes detection and relation classification mechanisms for the fully-supervised training scenario, and potentially achieve better performance. However, for investigating the fundamental difference coming from either using supervised and self-supervised training strategies, it is best to use the same network architectures as much as possible.

\paragraph{Metric:}

To evaluate the quality of scene graph, the image-wise SGGen metric is often used \cite{lu_visual_2016, xu_scene_2017, yang_graph_2018}. The idea is to organize the prediction and ground truth graphs as two sets of triplets with the format of \textit{<object, relationship, object>}. Then triplet matching is performed to compute the conventional Recall@k metric. The main reason only the recall is used and not also precision in previous work is that the manual annotations are sparse and it is not possible to annotate all existing relations between objects in images. However in our dataset, since the images are created from a given graph, we are able to capture the complete graph information. Thus, the precision metric is also applied. In our evaluation protocol, like previous work \cite{lu_visual_2016, xu_scene_2017, yang_graph_2018}, we also consider the task as detection of triplets, and use F1-score with underlying metrics being precision and recall. 

Since our model might have multiple predictions for the same triplet, we pre-process the raw prediction and delete the redundant triplets before evaluation. The recall is computed without sorting the confidence and all the predicted triplets are used for the computation of precision. In this sense, our F-score metric containing precision and recall covers the previously used SGGen, which is essentially recall, and is more challenging in terms of triplet redundancy evaluation, which is essentially the precision.

\paragraph{Implementation details:} 

We train our approach and the baseline using PyTorch \cite{paszke_automatic_2017}, and we use Adam \cite{kingma_adam:_2014} optimizer with batch size $= 128$, $\beta_1 = 0.6$, and $\beta_2 = 0.9$ for the training of both models. We also train the network in each setting with different random seeds for 10 times, and calculate the mean and standard deviation for each metric. For the \textit{proposed} approach, we train with the initial learning rate 0.0005 for 30 epochs and reduce the learning rate by the factor of 10 for the last 10 epochs. As for the \textit{baseline} approach, we notice that providing the full ground truth will significantly simplify the task, which is expected. Thus, we train the baseline model with the initial learning rate 0.0003 for 15 epochs and reduce the learning rate by the factor of 10 for the last 5 epochs. We find that the sub-task for the node attention prediction is particularly easy to train and over-fits quickly. Also, the supervision quality for the adjacency matrix is dependent on the quality of the predicted nodes, which is used for generating the temporary ground truth on the fly. To solve these issues for the fully-supervised baseline, for the first two epochs we train the node attention module and later we fix it and only train the relation classification module.

\subsection{Results}\label{sect_results}

\paragraph{Self-reconstruction \vs full supervision:}
We compare the performances of the proposed self-supervised approach and the fully-supervised baseline. The results are presented in Table~\ref{tab_main_qiantitative} and Figure~\ref{fig_qualitative}. 

Our unsupervised approach achieves a comparable performance as the baseline, \ie 67.9 and 61.3 respectively, in terms of F1-score. This shows that, even without any external explicit ground truth as a training target, the self-reconstruction is sufficient to provide supervision for the task of graph prediction. Unlike many other encoder-decoder approaches that process the data in the same format, \eg image-feature-image translation, in our case, the information in the bottleneck is a conventional graph with random node order and the corresponding adjacency matrix. By creating differentiable transformation modules, as introduced in Section~\ref{sect_method}, an image-graph-image auto-encoding framework is able to regress the graph information in the canonical formation automatically. As discussed before, note that we do not claim that our approach is absolutely better than the baseline in terms of the performance. One can optimize the design of the baseline in many aspects and achieve performance improvements. However, we opt to keep the network design the same to gain more meaningful insights into our self-supervised approach.

It must be clarified that the unsupervised approach contains several limitations that the supervised counterpart does not share. First, even though we use a simple dataset, it is less efficient for the network to learn under the self-supervised setting, which can be noticed by the number of training epochs used by two approaches. Second, if the task is more complicated and challenging, the self-supervised learning would be less effective or even fail, compared to the fully-supervised approach. Third, during the experiments we notice that the randomness has a larger impact on our self-supervised approach: although F1-scores are with similar standard deviation, the recalls have noticeable larger variance. With a very small probability, the network fails to reconstruct the full image, which is not likely to happen in a fully-supervised training scenario. 

\begin{table}
	\caption{Quantitative results of our approach and supervised baseline}
	\label{tab_main_qiantitative}
	\centering
	\begin{tabular}{lllll}
		\toprule
		Method     & Supervision & Precision & Recall & F1-score \\
		\midrule
		Ours & Image reconstruction  & \textbf{57.7}$\pm$2.7  & \textbf{84.1}$\pm$12.6 & \textbf{67.9}$\pm$5.4\\
		Baseline & Full (nodes \& edges) &  54.0$\pm$5.9 & 70.9$\pm$5.9 & 61.3$\pm$5.9 \\
		\bottomrule
	\end{tabular}
\end{table}

%
%
%

\paragraph{Effect of maximum number of nodes:}

In the previous main experiment we set the pre-defined maximum number of nodes to 4, according to the property of the dataset. In this experiment, we study the performance change when the defined maximum number of nodes is redundant even for the most complicated rectangle shape, which is the typical setting for real tasks. Table~\ref{tab_num_of_nodes} presents the quantitative results of different values for the maximum number of nodes. 

From the table, it can be concluded that the redundancy of maximum number of nodes will result in the performance drop of the triplets matching: 54.6, 49.9, and 41.7 F1-score when the number of nodes is 5, 6, and 8, respectively. This is mainly because, when the network has extra chances to reconstruct the image, the adjacency prediction tends to be conservative. For example, to reconstruct a triplet in an image, if there are two extra triplets for the network to predict, the network will tend to generate three triplets with the confidence of connectivity being 1/3. This will result in a reconstructed image that aligns with the input image due to triplets overlaying, but none of these triplets is correctly classified as connected. 

However, note that when the redundancy is limited, \eg 4 and 5,  the performance can remain at the acceptable level. Thus, in the real case that the size of the graph is unknown in a dataset without any graphs labels, as long as the maximum number of nodes is not extremely large, \eg 100\% extra nodes or more, one can expect to still have a decent graph quality.

\begin{table}
	\caption{Quantitative results of our approach with different defined maximum number of nodes}
	\label{tab_num_of_nodes}
	\centering
	\begin{tabular}{clll}
		\toprule
		Number of nodes & Precision & Recall & F1-score \\
		\midrule	
		4   & \textbf{57.7}$\pm$2.7  & \textbf{84.1}$\pm$12.6 & \textbf{67.9}$\pm$5.4\\
		5   & 46.1$\pm$3.5 & 68.9$\pm$14.9 & 54.6$\pm$6.9 \\
		6   & 43.2$\pm$6.7 & 62.0$\pm$12.9 & 49.9$\pm$5.9 \\
		8   & 31.4$\pm$2.9 & 62.3$\pm$5.5 & 41.7$\pm$2.8 \\
		\bottomrule
	\end{tabular}
\end{table}

\paragraph{Loss ablations:}
\label{sect_result_loss_ablations}

\begin{table}
	\caption{The results of our approaches trained with different losses.}
	\label{tab_loss_ablation}
	\centering
	\begin{tabular}{ccc|l}
		\toprule
		& \multicolumn{2}{c|}{$\mathcal{L}_{\text{main}}$ on } & \\
		$\mathcal{L}_{\text{aux}}$ & Refined image & Coarse image &  F1-score \\
		\midrule
		Yes & MS-SSIM & - &\textbf{67.9}$\pm$5.4 \\
		No & MS-SSIM & - &63.2$\pm$8.2 \\
		Yes & SSIM & - & 50.3$\pm$17.7 \\
		Yes & -   & MS-SSIM & 51.7$\pm$7.1 \\
		Yes & -   & SSIM & 12.1$\pm$9.7 \\
		\bottomrule
	\end{tabular}
\end{table}

We also perform the ablation study of different loss settings, to verify the design choice of the training loss. Five loss settings are tested, with their performances listed in Table~\ref{tab_loss_ablation}. For the sake of simplicity we only show F1-score in this experiment, since the precision and recall are highly correlated to it.

Our default setting (first row in Table~\ref{tab_loss_ablation}) exhibits the best performance, \ie 67.9 F1-score, and significantly outperforms the alternatives. If the auxiliary loss is disabled, one can observe a performance drop by 4.7 F1-score and a increase of standard deviation by 2.8. This validates the contribution of the auxiliary loss: it stabilizes the training procedure and improves the node attention quality without requiring human annotations.

When replacing the multi-scale structural similarity index measure (MS-SSIM) with regular SSIM loss (third row), a performance drop can be observed with a margin larger than 17 in F1-score. This is mainly because the structural similarity at multiple scales can better measure the graph difference represented by images and ignore the pixel-level image differences at higher resolutions. Instead of applying the losses after the refinement sub-network, we also apply them on the coarse reconstructed image directly provided by the differentiable CG module (fourth and fifth row). The results are 51.7 and 12.1 F1-score when using MS-SSIM and SSIM, respectively. The performance degradation is mainly due to the domain gap between the training images and online generated images from the differentiable CG module. This domain gap can also be observed in Figure \ref{fig_qualitative}. In this case, the supervision is not passed through the refinement network, thus the domain transformation is not performed. Since the (MS-)SSIM measures pixel-level similarity between two images, the domain gap will result in additional noise during the loss computation and thus inhibit the image reconstruction task which should be domain-invariant and only focus on the graph structure itself. This experiment also shows the importance of the refinement network and verifies that it can perform the domain adaption task via self-supervised training.

\section{Conclusion}

In this work, we propose a novel neural network that can learn to estimate graph information, \ie node and adjacency matrices, from image content without the need for manually annotated ground truth. At its core, the node and adjacency matrices are self-learned by properly designing the network architecture of the encoder, aligning it with a decoder based on differentiable image drawing techniques, and training this end-to-end differentiable system on basis of image reconstruction loss. In terms of the commonly used triplet matching metric, our approach achieves a performance comparable to the fully-supervised baseline. Although the current unsupervised approach is limited to line drawings of simple shapes and has certain limitations related to its task complexity and training stability, we believe this approach is an important stepping stone towards self-supervised learning of the image to graph translation task for more complex imagery.

\bibliographystyle{plainnat}
\bibliography{Library.bib}

\end{document}